\documentclass[letterpaper, 10 pt, conference]{ieeeconf}   

\IEEEoverridecommandlockouts                              

\makeatletter
\let\NAT@parse\undefined
\makeatother

\usepackage[pdftex]{graphicx}
\usepackage{amsmath} 
\usepackage{amssymb}  
\usepackage{subfigure}
\usepackage{multirow}
\usepackage{array,booktabs}
\usepackage{diagbox}
\usepackage{balance}
\usepackage{mathtools}
\usepackage{subfigure}
\usepackage{verbatim}
\usepackage[table]{xcolor}

\usepackage[numbers]{natbib}

\usepackage{irap_SIunits}
\usepackage{irap_acronyms}
\usepackage{irap_math}
\usepackage{irap_misc}

\usepackage{soul,color}

\usepackage{multirow}
\usepackage{subfigure}
\usepackage{soul,color}
\newcommand{\bl}[1]{{\textcolor{black}{#1}}}
\newcommand{\B}[1]{{\textbf{#1}}}

\usepackage{caption}

\title{\LARGE \bf
PrimA6D: Rotational Primitive Reconstruction\\for Enhanced and Robust 6D Pose Estimation
}

\author{Myung-Hwan Jeon${}^{1}$ and Ayoung Kim${}^{2*}$
\thanks{$^{1}$M. Jeon is with Department of the Robotics Program
			  KAIST, Daejeon, S. Korea \texttt{myunghwan.jeon@kaist.ac.kr}}%
\thanks{$^{2}$A. Kim is with the Department of Civil and Environmental Engineering,
        KAIST, Daejeon, S. Korea \texttt{ayoungk@kaist.ac.kr}}%
\thanks{This work was supported by MOLIT, S. Korea (20TSRD-B151228-01).}
}

\begin{document}

\maketitle
\thispagestyle{empty}
\pagestyle{empty}

\begin{abstract}

In this paper, we introduce a rotational primitive prediction based 6D object pose estimation using a single image as an input. We solve for the 6D object pose of a known object relative to the camera using a single image with occlusion. Many recent \ac{SOTA} two-step approaches have exploited image keypoints extraction followed by PnP regression for pose estimation. Instead of relying on bounding box or keypoints on the object, we propose to learn orientation-induced primitive so as to achieve the pose estimation accuracy regardless of the object size. We leverage a \ac{VAE} to learn this underlying primitive and its associated keypoints. The keypoints inferred from the reconstructed primitive image are then used to regress the rotation using PnP. Lastly, we compute the translation in a separate localization module to complete the entire 6D pose estimation. When evaluated over public datasets, the proposed method yields a notable improvement over the LINEMOD,  the Occlusion LINEMOD, and the YCB-Video dataset. We further provide a synthetic-only trained case presenting comparable performance to the existing methods which require real images in the training phase.

\end{abstract}


\section{Introduction}
\label{sec:intro}

The 6D object pose estimation intends to recover the orientation and translation of the object relative to the camera.
When solving 6D pose estimation, reaching accurate, fast and robust solutions has been a key element of augmented reality \cite{pvnet-2019-cvpr}, robotic tasks \cite{billings2018silhonet, periyasamy2018robust, wang2019densefusion}, and autonomous vehicles \cite{manhardt2019roi}. Despite the widespread use of RGB-D cameras, single image-based pose estimation is still essential when an RGB-D camera is not applicable.

Recent advances in deep learning-based object pose estimation has been groundbreaking. Direct end-to-end 6D pose regression suffered from the non-linearity of the rotation, and studies tackled this issue through orientation representation \cite{saxena-2009-icra} or combining \ac{PnP} and \ac{RANSAC}. The latter approaches reported current \acf{SOTA} performance by leveraging a two-step approach for pose estimation \cite{pvnet-2019-cvpr, brachmann2017dsac, learning-less-2017}. These approaches extract keypoints from the RGB image using deep learning and compute the 6D pose of the object using 2D-3D correspondence matching, such as the \ac{PnP} algorithm.



\begin{figure}[!t]
	\centering
	\includegraphics[width=0.9\columnwidth]{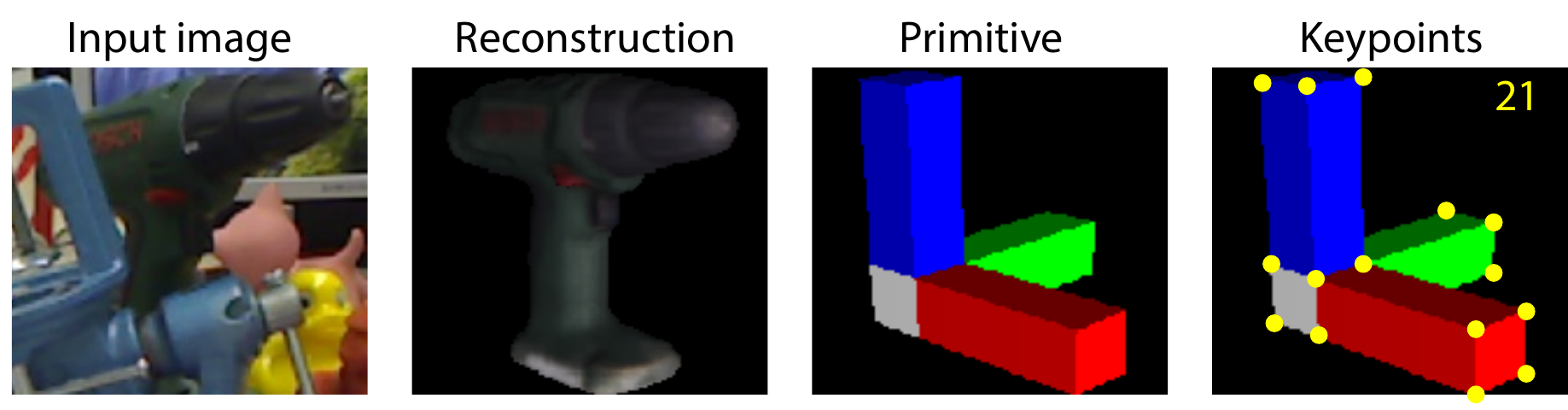}

	\caption{We train VAE to reconstruct objects and the associated primitive followed by the keypoints extraction over the inferred primitive. Keypoints are extracted from each corner plus the center, the same as the object center (i.e., 21).}

	\label{fig:oveview}
	\vspace{-5mm}
\end{figure}

We were also partially motivated by the finding in \cite{sundermeyer2018implicit} that an \ac{AAE} could implicitly represent a rotation when training an \ac{AAE} from a 3D CAD. The main advantage of this approach is that there is no need for manual annotation to prepare ground pose labeled training images. While \cite{sundermeyer2018implicit} only solved for the orientation retrieval given a discretized orientation set, we reconstruct a primitive from which keypoints are extracted under occlusion and complete a full orientation regression.

This paper solves for the direct 6D pose regression by introducing a novel primitive descriptor in \figref{fig:oveview}. The proposed solution, primitive associated 6D (PrimA6D) pose estimation, combines a direct and holistic understanding of an image with keypoint based PnP for the final orientation regression. More specifically, we newly introduce the primitive decoder in addition to the \ac{VAE} to increase the discriminability of the orientation inference. Via this step, like the aforementioned studies, we aim to find reliable keypoints in an image for pose regression extracting from the reconstructed primitives. Differing from previous methods, our method presents the following contributions.

\begin{figure*}[!t]
	\centering
	\includegraphics[width=0.95\textwidth]{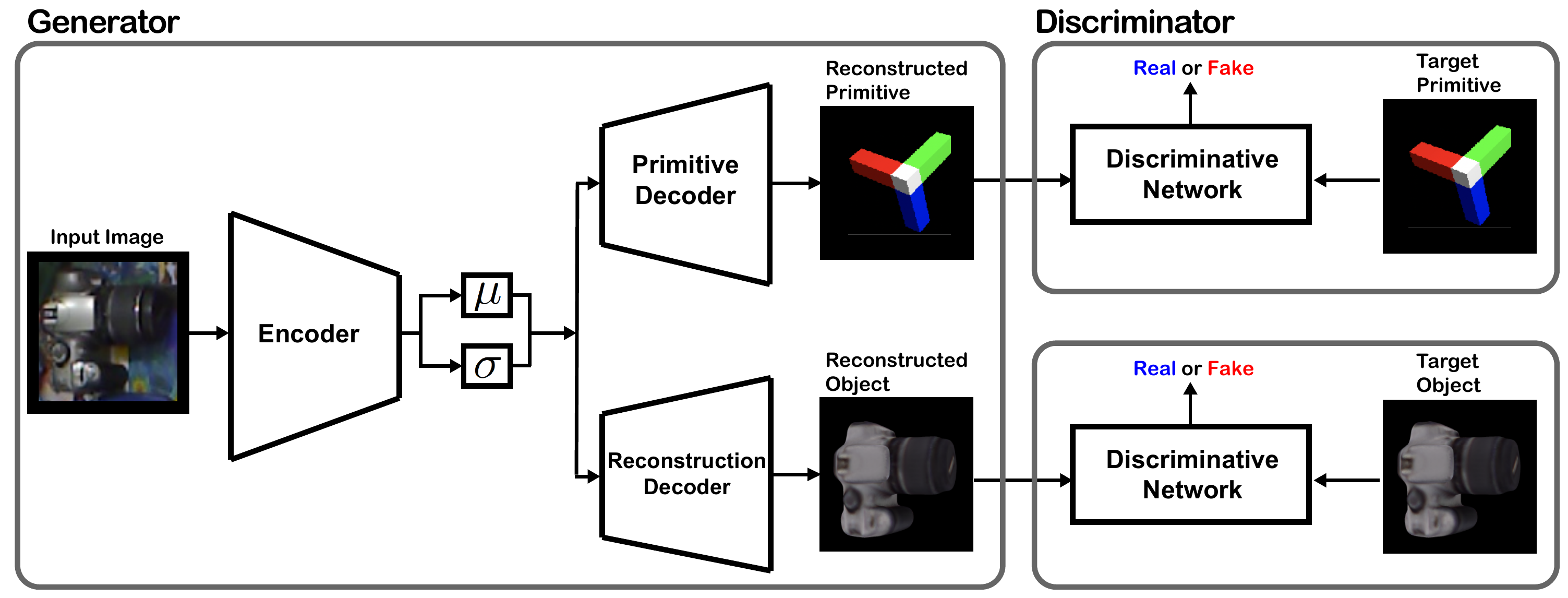}

	\caption{The reconstruction stage of PrimA6D. In the reconstruction stage, an encoder using ResNext50 \cite{DBLP:journals/corr/XieGDTH16} and an reconstruction decoder consisted of fully deconvolutional layers form the \ac{VAE}. We reconstruct both object and primitive in a separate decoder. To increase the discriminability, we additionally introduce adversarial loss and train using \ac{GAN}.}

	\label{fig:network}
	\vspace{-4mm}
\end{figure*}

\begin{itemize}

	\item We propose a novel 6D object pose estimation network called PrimA6D, which introduces a rotation primitive reconstruction and its associated keypoints to enhance the orientation inference.

	\item The proposed estimation scheme mitigates the manual effort of preparing label annotated images for training. The proposed primitive is straightforward to generate given rotation in 3D CAD; the trained \ac{VAE} allows us to learn from synthetic images via domain randomization robust to occlusion and symmetry.

	\item We verify the meaningful performance improvement of 6D object pose estimation compared to the existing \ac{SOTA} 6D pose regression methods. The proposed method improved estimation over LINEMOD as 97.62\% (PVNet 86.27\%), Occlusion LINEMOD as 59.77\% (PVNet 40.77\%), and YCB-Video as 94.43\% (PVNet 73.78\%) in terms of ADD(-S).

	\item The proposed method is more generalizable than existing methods showing less performance variance between datasets. This superior generalization capability is essential for robotic manipulation and automation tasks.

\end{itemize}

\section{Related Works}
\label{sec:related}



PnP based approaches establish 2D-3D correspondences from keypoints or patches and solve for the pose from these correspondences. Keypoint-based approaches define specific features to infer the pose from an image, such as the corner points of 3D bounding-box. These approaches chiefly involve two steps: first, the 2D keypoint of an object in the image is computed, and then an object pose is regressed using the \ac{PnP} algorithm. In BB8 \cite{rad2017bb8}, the authors predicted the 2D projection corner points of the object's 3D bounding box. Then, they computed the object pose from 2D-3D correspondences using \ac{PnP} algorithm. Similarly, \cite{tekin2018real} additionally added the central point of object to the keypoints of BB8 \cite{rad2017bb8}. Unfortunately, \cite{rad2017bb8,tekin2018real} revealed the weakness to the occlusion that prevents defining the correct bounding box of an object. To alleviate this issue, PVNet \cite{pvnet-2019-cvpr} predicted the unit vectors that point to predefined keypoints for each pixel in an object. By computing the locations of these keypoints using unit vector voting, the authors substantially improved their accuracy. However, these keypoint-based approaches may suffer when an object lacks texture, whereas the proposed primitive can be robustly reconstructed even for small and textureless objects.


The part-based approaches extract parts of an object called patches for pose regression. This method is  frequently used with depth data. For instance, \cite{brachmann2016uncertainty} jointly predicted the object labels and object coordinates for every pixel, then computed the object pose from the predicted 2D-3D correspondence matching. In \cite{oberweger2018making}, the authors extracted heatmaps of multiple keypoints from local patches then computed the object pose using the \ac{PnP} algorithm. Depth data may not always be available and we propose using a single RGB as input. Recently, \cite{pix2pose} exploited \ac{AE} to generate an image representing 3D coordinate per each pixel. Together with the reconstructed image containing 3D information, the expected error per pixel was also estimated to sort meaningful 2D-3D correspondence.




Unlike the aforementioned approaches, which consist of steps, holistic approaches utilize the overall shape of the object appearing in an image to directly regress the object pose relative to the camera. In Deep-6DPose \cite{do2018deep}, the authors found the regions where the objects were located through the \ac{RPN}. Despite solving for both object detection and 6D pose estimation, their method was vulnerable to small and symmetrical objects since they relied on the detected regions. In SSD-6D \cite{kehl2017ssd}, the authors used the discrete 6D pose space to represent the discriminable viewpoint. SSD-6D is a \ac{SSD}-style pose estimation model that extends the work of \cite{liu2016ssd}. Their method had a faster inference time than other methods though it sacrificed accuracy.

Incorporating a mask has been widely adopted. For example, \cite{billings2018silhonet,xiang2017posecnn,periyasamy2018robust,wu2018real} utilized an mask of the object to solve the pose estimation task. The well-known PoseCNN \cite{xiang2017posecnn} required an \ac{ICP} refinement for better accuracy. Stemming from PoseCNN, \cite{wu2018real} enabled a real-time performance with RGB data. The authors of \cite{xiang2017posecnn,wu2018real} also reported meaningful results, but their works were sensitive to the issue of object occlusion. Tackling the occlusion issue, SilhoNet \cite{billings2018silhonet} constructed an occlusion-free mask even from the occluded region, through CNN. This type of mask level restoration enhanced the performance over occlusion, however, yielded low accuracy for symmetrical objects. Unlike these methods relying on masks, we exploit the reconstructed object to not only regress the translation but further tackle the occlusion issue.


All the abovementioned methods require true pose-labeled images for training. Another stream of studies has focused on leveraging latent code without the need for labeled training images via \ac{AAE} to solve this issue. Without requiring the pose label from object 6D pose estimation, \cite{sundermeyer2018implicit} utilized the latent code generated from the object recovery process using \ac{AE}. The object pose was determined by a similarity comparison with the predefined latent codebook for each pose, but the accuracy was dominated by the size of the codebook with a discretized orientation class. We have overcome this accuracy limitation and achieved substantially improved performance from a synthetic dataset.


\section{Method}
\label{sec:method}

\begin{figure}[!t]
	\centering
	\includegraphics[width=1\columnwidth]{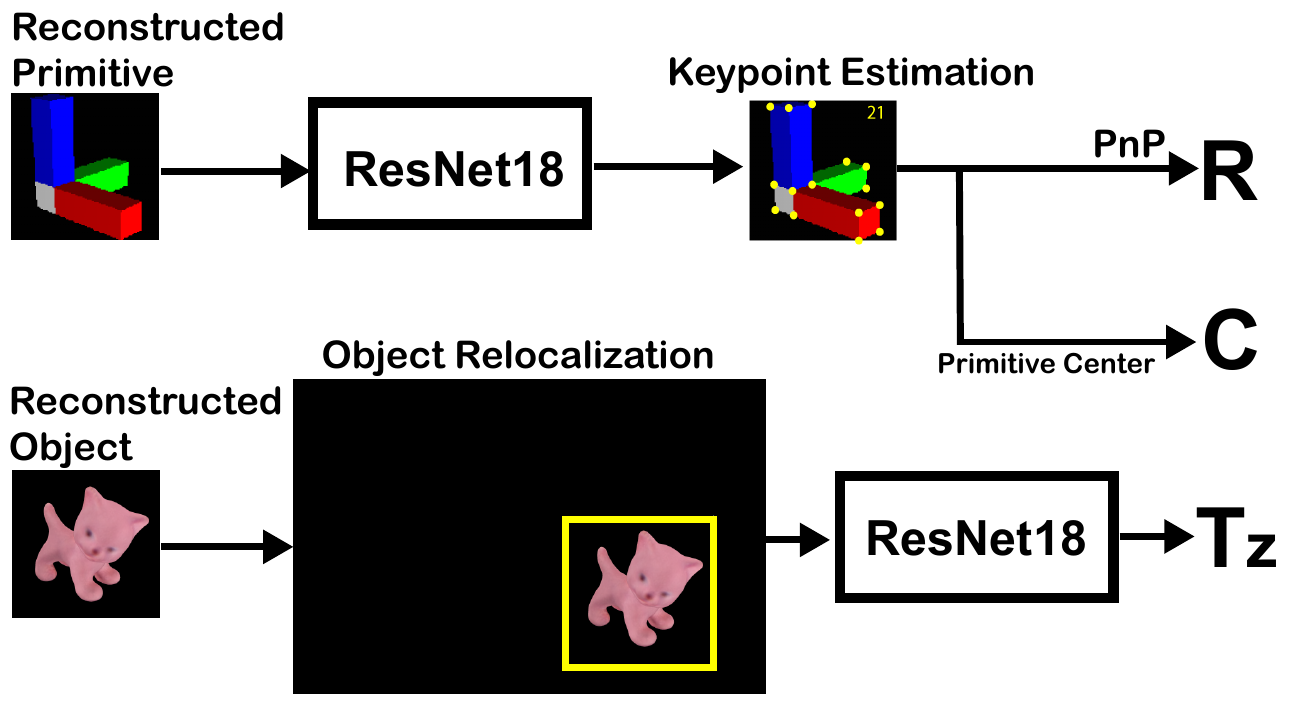}
  \caption{The rotation and translation inference. We use keypoints learned from the reconstructed primitive for the orientation estimation using PnP. In the translation inference stage, the network input is created using the output of the reconstruction decoder through object relocalization. }

	\label{fig:pose}
	\vspace{-1mm}
\end{figure}

Given an RGB image, we compute the 6D pose of an object relative to the camera in terms of the its relative transform ${\mathbf{T}}^{C}_{O}$ between the camera coordinate frame ${C}$ and the object coordinate frame ${O}$. The 6D pose is represented by quaternion and translation vectors. Similar to \cite{pix2pose} and \cite{sundermeyer2018implicit}, the proposed method can be accompanied with other detection modules such as M2Det \cite{DBLP:journals/corr/abs-1811-04533}. We use a cropped image using a bounding box detected from other methods as input to our network.



\subsection{Training Set Prepration}

As indicated earlier, our aim is at enabling 6D pose regression without requiring the real images. For each pair of a 3D CAD object and the associated camera pose, we prepare the training set consists of a bounding box, rendered primitive, the target keypoints (the 20 corner points and a center point), and rendered training image. The primitive images represent the rotation of the object having different colors per axis as in \figref{fig:oveview}.

\subsection{Object and Primitive Reconstruction}

Using the training data, the first phase of the proposed method consists of object and primitive reconstruction as shown in \figref{fig:network}. We train the \ac{VAE} having an encoder and two reconstruction decoders. This step is partially similar to \ac{AAE} in \cite{sundermeyer2018implicit}. In \cite{sundermeyer2018implicit}, the authors verified that a trained \ac{AAE} using a geometric augmentation technique learns the representation of the object's orientation and generates the latent code with data for the object orientation. We adopt the same technique in the training phase by rendering the training images with various poses of the object. The training images are augmented using the domain randomization technique \cite{mjeon-2019-cams} to bridge the gap between real and rendered images. Augmented images are fed into the encoder, and clean target images are obtained through the reconstruction decoder using four losses as follows. Differing from \cite{sundermeyer2018implicit}, our training strategy is to use \ac{GAN} by introducing a discriminator for the reconstructed object and primitive.

\subsubsection{Object Reconstruction Loss}

The first loss focuses on the reconstruction of an object by comparing the pixel-wise loss. To prevent overfitting in the reconstruction stage, we use the top-k pixel-wise L2 loss as follow.
\begin{equation}
	\label{eq:reconst_loss}
	\mathcal{L}_{O}=\frac{1}{K}\sum_{i=1}^{K} \ell_{[i]}(\left \| x-\hat{x} \right \|^2),
\end{equation}
where $\ell_{[i]}$ is a function that extracts the pixel with the $i^\text{th}$ largest error \cite{atkloss-2017-nips}. We use $K=128$ in this paper. In \eqref{eq:reconst_loss}, $x$ is the target image pixel and $\hat{x}$ is the predicted image pixel respectively. This initial latent code encompasses the orientation but it is not sufficiently discriminative for regression. To increase the discriminability, we introduce the rotational primitive decoder to embolden the orientation of the code.

\subsubsection{Primitive Reconstruction Loss}

The rendered primitive image becomes the prediction target of the primitive decoder. Using the color-coded axes in primitive image, we designed the color based axis-aligning loss function for the primitive decoder. When computing the loss for the channels, we consider a per-pixel intensity difference $d_c$ for each channels $c \in \mathcal{C} = [R,G,B]$.
\begin{eqnarray}
	\label{eq:primitive}
	d_{c} &=& x_{c}-\hat{x}_{c}\\
	S_{c} &=&\frac{1}{K}\sum_{i=1}^{K}\ell_{[i]}(\exp({\alpha \left | d_c \right |}) \cdot \left \|  d_c \right \|^{2}) \\
	C_{c} &=&\frac{1}{K}\sum_{i=1}^{K}\ell_{[i]}( \left \|  d_c \right \|^{2})\\
	\mathcal{L}_{P}&=&\sum_{c \in \mathcal{C}}S_{c} \exp \left( \frac{C_{c}}{\sum_{k \in \mathcal{C}}C_{k}} \right)
\end{eqnarray}
Here, the $S_{c}$ measures misalignment of each axis in terms of the pixel intensity difference per channel. The $C_{c}$ measures the overall misalignment per channel and discerns between channels by penalizing the channel showing a larger error when the primitive reconstruction loss $L_P$ is constructed. The $\alpha$ is a weight constant ($\alpha=5$ in this paper).

\subsubsection{Overall VAE Loss}

Using these two reconstruction losses with \ac{KL} divergence loss, we train the \ac{VAE} to generate both an object image and a primitive image. By adopting \ac{VAE}, the network encodes input image $x$ to the parameters of a Gaussian distribution $q(z|x)$. In doing so, we minimize the \ac{KL} divergence between $q(z|x)$ and $\mathcal{N}(0,I)$. The overall loss function for the reconstruction stack becomes as below.
\begin{equation}
	\label{eq:loss_kld}
	\mathcal{L}_{R} = \mathcal{L}_{O} + \mathcal{L}_{P} + D_{KL} \left( q(z|x)||\mathcal{N}(0,I) \right)
\end{equation}

\subsubsection{Adversarial Loss}

We further aim to improve the quality of the reconstruction using training strategy of \ac{GAN}. For instance, the reconstructed primitive sometimes becomes ambiguous as shown in \figref{fig:effect_gan} and may result in incorrect keypoints learning. Regarding this issue, introducing adversarial loss improved the primitive/object prediction accuracy and thereby the overall pose inference performance. For the target primitive/object image $X$ and the reconstructed image $\hat{X}$, we consider the following loss to train using \ac{GAN} \cite{goodfellow2014generative}.
\begin{equation}
	\label{eq:loss_adv}
\mathcal{L}_{A} = \mathbb{E}_{\hat{X}}[\log D(\hat{X})] + \mathbb{E}_{X}[\log (1-D(G(X)))].
\end{equation}
Here we use $D{(\cdot)}$ to distinguish if the predicted primitive/object is the target primitive/object from 3D model (true) or generated by $G{(\cdot)}$ (fake). Following the training strategy of \ac{GAN}, we solve for $G$ and $D$ by minimizing $\mathcal{L}^G_{A}=\mathbb{E}_{\hat{X}}[\log D(\hat{X})]$ and maximizing $\mathcal{L}^D_{A}=\mathbb{E}_{X}[\log (1-D(G(X)))]$.

\begin{figure}[!t]
	\centering
	\includegraphics[width=0.8\columnwidth]{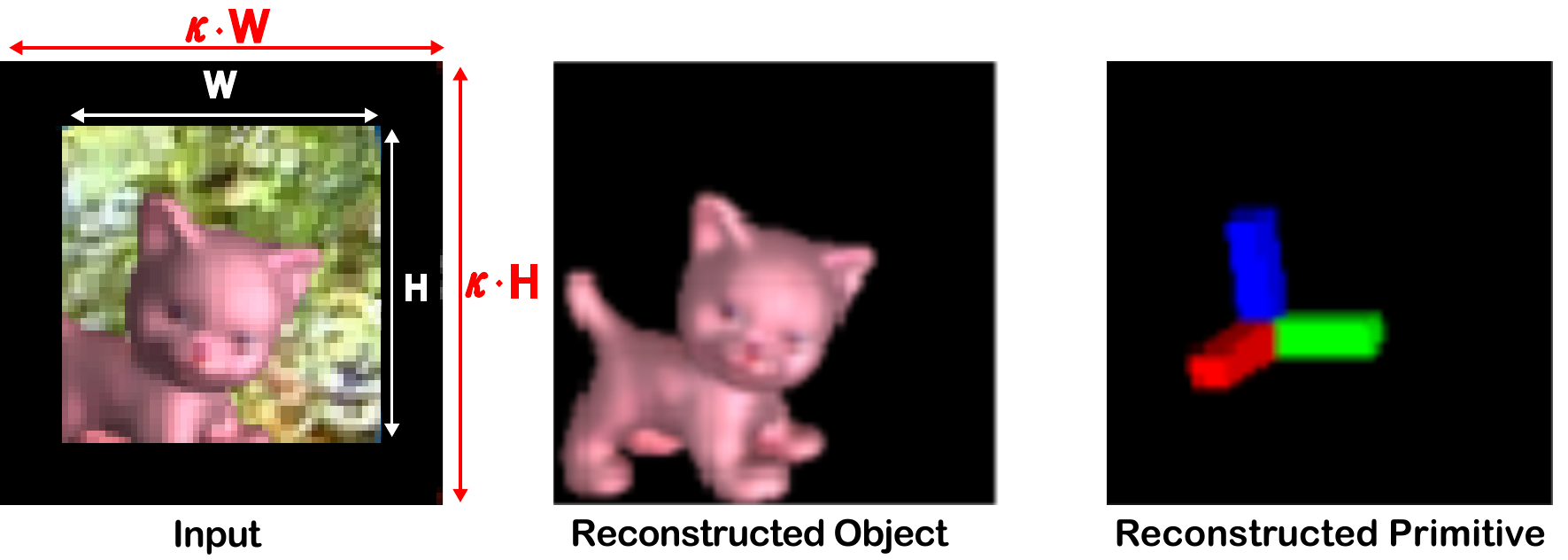}
	\caption{The input image of the reconstruction network is ($\kappa$=1.3) larger than the actual bounding-box allowing complete object reconstruction even for inaccurate bounding-box.}

	\label{fig:synthetic_dataset}
	\vspace{-4mm}
\end{figure}

\subsection{Orientation Regression via PnP}

From the reconstructed primitive we estimate the rotation first by learning the keypoints from it. Using generated keypoints in the training set preparation and via pixel-wise L2 loss, we adopt ResNet18 \cite{resnet} and extract the keypoints of the reconstructed primitive using the loss as
\begin{equation}
	\label{eq:pnp_R}
	\mathcal{L}_{rot}=\sum\left \| x_k-\hat{x_k} \right \|^2.
\end{equation}
Since the keypoints consists of corners and the center point of the object which is the same as the primitive center, we utilize this estimated center point, $C=x_c = (u_c,v_c)$ for the translation estimation.

\subsection{Translation Regression}

\begin{table*}[!t]
\centering
\resizebox{\linewidth}{!}{
\begin{tabular}{c|cc|ccccccc|ccc}
  \hline
  \multirow{3}{*}{methods} & \multicolumn{2}{c|}{\multirow{2}{*}{Holistic Approach}}          & \multicolumn{9}{c}{PnP based Approach} \\ \cline{4-13}
                           &                                 &                                & \multicolumn{7}{c}{w/o refinement} & \multicolumn{3}{|c}{\bl{w/ refinement}}\\ \cline{2-13}
                           & PoseCNN \cite{xiang2017posecnn} & Deep-6D Pose \cite{do2018deep} &  BB8 \cite{rad2017bb8} & Tekin \cite{tekin2018real} & Pix2Pose \cite{pix2pose} & DPOD \cite{zakharov2019dpod} &  PVNet \cite{pvnet-2019-cvpr} & PrimA6D-S & PrimA6D-SR &  BB8 & Tekin & HybridPose \cite{song2020hybridpose}\\ \hline \hline
                           ape$^\dagger$& -     & 38.8 & 27.9        & 21.62       & 58.1   & 53.28     & 43.62       & 66.5       & \B{90.21}   & 40.4        & 65        &  77.6  \\ \hline
                           benchvise    & -     & 71.2 & 62          & 81.8        & 91     & 95.34     & \B{99.9}    & 97.11       & 99.75       & 91.8        & 80       & 99.6   \\ \hline
                           cam          & -     & 52.5 & 40.1        & 36.57       & 60.9    & 90.36    & 86.86       & 91.5       & \B{98}   & 55.7        & 78          & 95.9\\ \hline
                           can          & -     & 86.1 & 48.1        & 68.8        & 84.4   & 94.10     & 95.47       & 93.89       & \B{98.74}   & 64.1        & 86      & 93.6    \\ \hline
                           cat$^\dagger$& -     & 66.2 & 45.2        & 41.82       & 65    & 60.38      & 79.34       & 88.21       & \B{99.06}   & 62.6        & 70      & 93.5    \\ \hline
                           driller      & -     & 82.3 & 58.6        & 63.51       & 76.3   & 97.72     & 96.43       & 98.14       & \B{99.57}   & 74.4        & 73      & 97.2    \\ \hline
                          duck$^\dagger$& -     & 32.5 & 32.8        & 27.23       & 43.8   & 66.01     & 52.58       & 72.56       & \B{90.82}   & 44.3        & 66       & 87.0   \\ \hline
                           eggbox*      & -     & 79.4 & 40          & 69.58       & 96.8  & 99.72      & 99.15   & 91.93       & 99.52        & 57.8        & \B{100}   & 99.6      \\ \hline
                           glue*        & -     & 63.7 & 27          & 80.02       & 79.4   & 93.83     & 95.66       & 91.14       & 99.75   & 41.2        & \B{100}     & 98.7    \\ \hline
                   holepuncher$^\dagger$& -     & 56.4 & 42.4        & 42.63       & 74.8   & 65.83     & 81.92       & 68.71       & \B{95.31}   & 67.2        & 49    & 92.5      \\ \hline
                           iron         & -     & 65.1 & 67          & 74.97       & 83.4   & \B{99.80}     & 98.88   & 97.3       & 99.56          & 84.7        & 78    & 98.1     \\ \hline
                           lamp         & -     & 89.4 & 39.9        & 71.11       & 82    & 88.11      & \B{99.33}   & 95.59       & 99.18       & 76.5        & 73       & 96.9   \\ \hline
                           phone        & -     & 65   & 35.2        & 47.74       & 45    & 74.24      & 92.41       & 89.38       & \B{99.59}   & 54          & 79      & 98.3    \\ \hline \hline
                           average      & 55.95 & 65.2 & 43.55       & 55.95       & 72.38  & 82.98     & 86.27       & 87.84       & \B{97.62}   & 62.67       & 76.69    & 94.5   \\ \hline
  \end{tabular}
  }
  \vspace{0.1mm}
  \caption{The accuracy in terms of the ADD(-S) metric for our method and baseline methods on the LINEMOD dataset. Symmetrical objects are marked with an asterisk (*) and small objects are marked with $^\dagger$. PoseCNN \cite{xiang2017posecnn} only provided the average.}
  \label{tab:linemod_add}
\end{table*}

\begin{table*}[!t]
\centering
\resizebox{\linewidth}{!}{
\begin{tabular}{c|cc|ccc|c}
  \hline
  method & \multicolumn{5}{c|}{w/o refinement} & w/ refinement \\ \hline
         & BB8 & Tekin & PVNet & PrimA6D-S & PrimA6D-SR &  BB8 \\ \hline \hline
       ape$^\dagger$ & 95.3        & 92.1        & \B{99.23}   & 97.24       & 99.02       & 96.6        \\ \hline
         benchvise   & 80          & 95.06       & \B{99.81}   & 95.3        & 98.92       & 90.1        \\ \hline
         cam         & 80.9        & 93.24       & 99.21   	   & 97       	 & \B{99.75}   & 86          \\ \hline
         can         & 84.1        & 97.44       & \B{99.9}    & 93.64       & 99.24       & 91.2        \\ \hline
       cat$^\dagger$ & 97          & 97.41       & 99.3    	   & 95.08       & \B{99.83}   & 98.8        \\ \hline
         driller     & 74.1        & 79.41       & 96.92   	   & 97.22       & \B{98.9}    & 80.9        \\ \hline
      duck$^\dagger$ & 81.2        & 94.65       & 98.02   	   & 96.65       & \B{98.8}    & 92.2        \\ \hline
         eggbox*     & 87.9        & 90.33       & \B{99.34}   & 86.59       & 98.32       & 91          \\ \hline
         glue*       & 89          & 96.53       & 98.45       & 91.22       & \B{99.26}   & 92.3        \\ \hline
holepuncher$^\dagger$& 90.5        & 92.86       & \B{100}     & 81     	 & 97.41       & 95.3        \\ \hline
         iron        & 78.9        & 82.94       & \B{99.18}   & 84.89       & 98.58       & 84.8        \\ \hline
         lamp        & 74.4        & 76.81       & 98.27   	   & 78.72       & \B{98.77}   & 75.8        \\ \hline
         phone       & 77.6        & 86.07       & \B{99.42}   & 93.56       & 99.11       & 85.3        \\ \hline \hline
         average     & 83.92       & 90.37       & \B{99.00}   & 91.39       & 98.92       & 89.25       \\ \hline
  \end{tabular}
  \hspace{5mm}
  \begin{tabular}{c|cccc|cccc}
    \hline
    method & \multicolumn{4}{c|}{Rotation MAE [deg]} & \multicolumn{4}{c}{Translation MAE [mm]} \\ \hline
           & Tekin & PVNet & PrimA6D-S & PrimA6D-SR & Tekin & PVNet & PrimA6D-S & PrimA6D-SR \\ \hline \hline
           ape$^\dagger$& 5.69 & 6.25 & 6.35 & \B{3.63} & 29.32 & 47.70 & 10.58 & \B{5.72}  \\ \hline
           benchvise    & 3.14 & \B{1.43} & 3.32 & 1.96 & 18.86 & 6.97  & 6.94 & \B{3.81}  \\ \hline
           cam          & 3.91 & \B{1.75} & 3.05 & 1.83 & 33.47 & 9.74  & 7.31 & \B{4.29} \\ \hline
           can          & 2.89 & \B{1.48} & 3.14 & 2.01 & 16.87 & 7.93  & 8.47 & \B{5.55}  \\ \hline
           cat$^\dagger$& 4.79 & \B{1.88} & 3.91 & 2.27 & 22.61 & 10.56 & 8.27 & \B{3.35}  \\ \hline
           driller      & 3.42 & 2.42 & 2.30 & \B{1.53} & 19.72 & 16.67 & 4.46 & \B{3.20} \\ \hline
         duck$^\dagger$ & 6.39 & 3.00 & 4.87 & \B{2.72} & 28.52 & 15.51 & 8.83 & \B{4.84}  \\ \hline           
   holepuncher$^\dagger$& 4.77 & \B{1.99} & 7.82 & 3.29 & 31.37 & 9.70  & 15.99 & \B{4.80}  \\ \hline
           iron         & 6.21 & 2.07 & 3.27 & \B{1.74} & 27.87 & 10.95 & 9.41 & \B{4.57}  \\ \hline
           lamp         & 4.24 & \B{1.50} & 4.11 & 1.62 & 26.78 & 7.90  & 9.64 & \B{4.35} \\ \hline
           phone        & -    & \B{1.82} & 3.69 & 1.98 & -     & 9.90  & 9.29 & \B{4.46}  \\ \hline \hline
           average      & 4.55 & 2.33 & 4.17 & \B{2.23} & 25.54 & 13.96 & 9.02 & \B{4.45} \\  \hline

    \end{tabular}
    }
  \vspace{0.1mm}
  \caption{The accuracy in terms of the 2D projection error (left) and mean absolute error (MAE) (right) for our method and baseline methods on the LINEMOD dataset. Symmetrical objects are marked with an asterisk (*) and small objects are marked with $^\dagger$. Using the provided network and weight by authors, we evaluate mean absolute error. Tekin did not provide weight for the phone model. On evaluation in terms of MAE, the symmetric object is excluded because evaluation for the symmetric object requires a shape comparison rather than a specific pose comparison.}  
  \label{tab:linemod_2dp}
\end{table*}


Given the estimated orientation, we then infer the 3D translation by computing the scene depth $T_{z}$ of the object center (i.e., primitive center). The reconstructed image is relocated to the original image using the bounding box of the object. This process is called \textit{object relocalization} (\figref{fig:pose}).

As the sample illustration in \figref{fig:synthetic_dataset} shows, the reconstruction occurs \textit{without} assuming the bounding box is centered and \bl{thereby is robust to inaccurate object detection.} The reconstruction phase handles the inaccurate bounding box location by properly reconstructing the object even at the off-centered position. The inflation scaling factor $\kappa$ was chosen to handle the detected bounding box with \ac{IoU} \bl{0.75}. A small $\kappa$ fails to handle inaccuracy in objection detection phase, and a large $\kappa$ might deteriorate the performance with increased empty space near borders. Because this re-localization replaces an object with a reconstructed object, the translation can be regressed over occlusion-free objects. From this relocalized full image, we again use ResNet18 to regress the scene depth $T_{z}$ to the object center.

Using $T_{z}$, the bounding-box, and the camera intrinsic matrix $K$, we compute the $T_{x}$ and $T_{y}$ components of translation as follows from the estimated object center $(u_c, v_c)$. The focal length $f$ and principal point $(u_p,v_p)$ are obtained from the calibration matrix while the bounding box is known when preparing the training data.
\begin{eqnarray}
	\label{eq:trans}
	u_c=f_{u}\frac{T_x}{T_z}+u_p \quad\text{and}\quad
	u_v=f_{v}\frac{T_y}{T_z}+v_p.
\end{eqnarray}
%





\section{Experiment}
\label{sec:experiment}

In this section, we evaluate the proposed method using three different datasets.

\begin{table*}[!t]
\centering
\resizebox{\linewidth}{!}{
\begin{tabular}{c|ccccc|ccc|c}
\hline
\multirow{2}{*}{methods} & \multicolumn{8}{c|}{\multirow{1}{*}{w/o refinement}}          & \multicolumn{1}{c}{w/ refinement} \\ \cline{2-10}
       & Tekin & PoseCNN & Oberweger\cite{oberweger2018making} & Pix2Pose & PVNet & PrimA6D-S & PrimA6D-SR  & w/o GAN & HybridPose\\ \hline \hline
	 ape$^\dagger$& 2.48  & 9.6   & 17.6  & 22    & 15.81 & 28.71 & 37.35 &2.75 &\B{53.3}\\ \hline
	 can          & 17.48 & 45.2  & 53.9  & 44.7  & 63.3  & 61.88 & 67.19 &5.47 &\B{86.5}\\ \hline
	 cat$^\dagger$& 0.67  & 0.93  & 3.31  & 22.7  & 16.68 & 40.43 & 44.65 &1.68 &\B{73.4}\\ \hline
	duck$^\dagger$& 1.14  & 19.6  & 19.2  & 15    & 25.24  &39.28 & 50.65 &10.87 &\B{92.8}\\ \hline
	 driller      & 7.66  & 41.4  & 62.4  & 44.7  & 65.65  & 70.42 & \B{74.87} &9.71\ &62.8\\ \hline
	 eggbox*      & -     & 22    & 25.9  & 25.2  & 50.17 & 54.31 & 66.01 &7.02 &\B{95.3}\\ \hline
	 glue*        & 10.08 & 38.5  & 39.6  & 32.4  & 49.62  & 71.03 & 73.02 &7.72 &\B{92.5}\\ \hline
	holepuncher$^\dagger$& 5.45  & 22.1  & 21.3  & 49.5   & 12.41 & 46.19 & 64.38 &12.41 &\B{76.7}\\ \hline \hline
	 average      & 6.42  & 24.92 & 30.40 & 32.00 & 40.77  & 51.53 & 59.77 &7.20 &\B{79.2}\\ \hline
\end{tabular}
\hspace{5mm}
\begin{tabular}{c|cccc|cc}
	\hline
	methods & Tekin & PoseCNN & Oberweger & PVNet &  PrimA6D-S & PrimA6D-SR \\ \hline \hline
	ape$^\dagger$& 7.01 & 34.6  & \B{69.9}  & 69.14 &  56.58 & 62.64 \\ \hline
	can          & 11.2 & 15.1  & 82.6  & \B{86.09} &  69.26 & 76.8 \\ \hline
	cat$^\dagger$& 3.62 & 10.4  & 65.1  & \B{65.12} &  54.33 & 58.8 \\ \hline
	duck$^\dagger$& 5.07 & 31.8  & 61.4  & 61.44     &  64.74 & \B{67.36} \\ \hline
	driller      & 1.4  & 7.4   & \B{73.8}  & 73.06 &  59.55 & 61.28 \\ \hline
	eggbox*      & -    & 1.9   & 13.1  & 8.43      &  8.53  & \B{16.48} \\ \hline
	glue*        & 4.7  & 13.8  & 54.9  & 55.37 &  51.83 & \B{57.38}  \\ \hline
	holepuncher$^\dagger$& 8.26 & 23.1  & 66.4  & \B{69.84}     &  50.82 & 67.76 \\ \hline \hline
	average      & 5.89 & 17.26 & 60.90 & \B{61.06} &  51.96 & 58.56 \\ \hline
\end{tabular}
}
	\vspace{0.1mm}
\caption{The accuracy in terms of the ADD(-S) metric (left) and 2D projection error (right) for our method and baseline methods on the Occlusion LINEMOD dataset. All values are imported from \cite{pvnet-2019-cvpr} except pix2pose and hybridpose. Symmetrical objects are marked with an asterisk (*) and small objects are marked with $^\dagger$.}
\label{tab:occ_add}
\end{table*}


\subsection{Dataset}

In total, three datasets were used for the evaluation: LINEMOD \cite{hinterstoisser2012model}, Occlusion LINEMOD \cite{hinterstoisser2012model}, and YCB-Video \cite{xiang2017posecnn}. The LINEMOD and Occlusion LINEMOD datasets are widely exploited benchmarks for 6D object pose estimation and cover various challenges, including occlusion and texture-less. The LINEMOD dataset comprises 13 objects with 1,312 rendered images on each for the training and about 1,200 images per object for the test. The Occlusion-LINEMOD dataset shares the training images with the LINEMOD dataset. For the test, they provide pose information for eight occluded objects. A recently released dataset, the YCB-Video dataset is composed of 21 high-quality 3D models, offering 92 annotated video sequences per frame. These video sequences include various lighting conditions, noise in the capture, and occlusion.

\subsection{Training Detail}

We sampled 50,000 object poses and rendered 50,000 training images per object together with their associated primitives. The rendered object images were further augmented using the method proposed by \cite{mjeon-2019-cams} using domain randomization. We train our network mainly from synthetic images and additionally train with 180 real images for three types of dataset.

All of the networks were trained with the Adam optimizer on Titan V. The ResNet backbone was initialized with pre-trained weights provided in PyTorch. All networks were trained with a batch size of 50 and a 0.0001 learning rate for 40 epochs.

\subsection{Evaluation Metric}

\textbf{(i) 2D Projection error metric} To evaluate the pose estimation in terms of the 2D projection error we also use the same metric as in \cite{pvnet-2019-cvpr} and measure the mean pixel distance between projection of the 3D model and the image pixel points. We also count the estimation is correct if the error is less than 5 pixels.

\textbf{(ii) 3D Projection error metric} Similar to \cite{xiang2017posecnn}, two metrics were selected for evaluation. We chose the average distance metric (ADD)~\cite{hinterstoisser2012model} as the evaluation metric, which computes the distance of transformed 3D model points between the ground truth pose and predicted pose using ADD(-S) metric.
%
%
When the average distance value is less than 10\% of the 3D model's diameter, the predicted 6D pose is considered to be correct. This ADD metric, however, is not suitable to use with symmetrical objects. For symmetrical objects, the average distance is computed between the closest points.
%

\textbf{(iii) Pose estimation metric} For the original LINEMOD dataset, we directly measure the \ac{MAE} for rotation and translation together with other metrics.

\begin{table*}[!t]
  \centering
  \resizebox{\linewidth}{!}{
  \begin{tabular}{c|ccccc}
    \hline
     methods & PoseCNN & Oberweger & PVNet & PrimA6D-S & PrimA6D-SR \\ \hline \hline
             002\_mater\_chef\_can  & 55.17 & 49.1  & 81.6  & 82.57 & \B{99.71}  \\ \hline
             003\_cracker\_box      & 52.9  & 83.6  & 80.5  & 75.7  & \B{99.44} \\ \hline
             004\_sugar\_box        & 68.3  & 82    & 84.9  & 97.94 & \B{99.74} \\ \hline
             005\_tomato\_soup\_can & 66.1  & \B{79.7}  & 78.2  & 73.51 & 78.39 \\ \hline
             006\_mustard\_bottle   & 80.8  & 91.4  & 88.3  & 94.42 & \B{99.96}\\ \hline
007\_ tuna\_fish\_can$^\dagger$     & 70.6  & 49.2  & 62.2  & 56.77 & \B{79.47}\\ \hline
           008\_pudding\_box        & 62.2  & 90.1  & 85.2  & 52.06 & \B{99.59}\\ \hline
           009\_gelatin\_box        & 74.8  & 93.6  & 88.7  & 94.88 & \B{99.59} \\ \hline
010\_potted\_meat\_can$^\dagger$    & 59.5  & 79    & 65.1  & 38.72 & \B{80.35}  \\ \hline
           011\_banana              & 72.1  & 51.9  & 51.8  & 74.14 & \B{87.01}\\ \hline
           019\_pitcher\_base       & 53.1  & 69.4  & 91.2  & 71.94 & \B{98.12} \\ \hline
           021\_bleach\_cleanser    & 50.2  & 76.1  & 74.8  & 50.75 & \B{90.04}\\ \hline
           024\_bowl*               & 69.8  & 76.9  & 89    & 96.84 & \B{98.79}\\ \hline
           025\_mug                 & 58.4  & 53.7  & 81.5  & 88.82 & \B{89.93}\\ \hline
           035\_power\_drill        & 55.2  & 82.7  & 83.4  & 92.62 & \B{97.67}\\ \hline
           036\_wood\_block*        & 61.8  & 55    & 71.5  & 95.68 & \B{100}  \\ \hline
           037\_scissors            & 35.3  & 65.9  & 54.8  & 63.89 & \B{97.85}\\ \hline
           040\_large\_marker       & 58.1  & 56.4  & 35.8  & 85.51 & \B{94.18}\\ \hline
           051\_large\_clamp*       & 50.1  & 67.5  & 66.3  & 97.72 & \B{98.89}\\ \hline
           052\_extra\_large\_clamp*& 46.5  & 53.9  & 53.9  & 88.25 & \B{98.12}\\ \hline
           061\_foam\_brick*        & 85.9  & 89    & 80.6  & 80.91 & \B{96.15}\\ \hline \hline
           average                  & 61.28 & 71.24 & 73.78 & 78.74 & \B{94.43}\\ \hline
  \end{tabular}
  \hspace{5mm}
  \begin{tabular}{c|ccccc}
    \hline
     methods & PoseCNN & Oberweger & PVNet & PrimA6D-S & PrimA6D-SR \\ \hline \hline
         002\_mater\_chef\_can   & 74.2 & 0.09  & 29.7  & 24.5  & \B{97.24} \\ \hline
         003\_cracker\_box       & 0.12 & 64.7  & 50.35 & 21.47 & \B{93.24} \\ \hline
         004\_sugar\_box         & 7.11 & 72.2  & 61.25 & 74.47 & \B{94.11} \\ \hline
         005\_tomato\_soup\_can  & 5.21 & 39.8  & 60.69 & 85.31 & \B{89.74}  \\ \hline
         006\_mustard\_bottle    & 6.44 & \B{87.7}  & 82.35 & 74.06 & 87.28 \\ \hline
   007\_tuna\_fish\_can$^\dagger$& 2.96 & 38.9  & 45.21 & 97.21 & \B{99.4} \\ \hline
         008\_pudding\_box       & 5.14 & 78    & 52.8  & 69.63 & \B{99.76} \\ \hline
         009\_gelatin\_box       & 15.8 & 94.8  & 94.85 & 52.12 & \B{97.78} \\ \hline
010\_potted\_meat\_can$^\dagger$ & 23.1 & 41.2  & 62.92 & 69.25 & \B{95.86} \\ \hline
         011\_banana             & 0.26 & 10.3  & 8.18  & 37.85 & \B{87.01} \\ \hline
         019\_pitcher\_base      & 0    & 5.4   & \B{79.3}  & 46.54 & 69.64 \\ \hline
         021\_bleach\_cleanser   & 1.16 & 23.2  & 37.51 & 58.61 & \B{67.11} \\ \hline
         024\_bowl*              & 4.43 & 26.1  & 33.99 & 26.59 & \B{56.13} \\ \hline
         025\_mug                & 0.78 & 29.2  & 52.98 & 89.75 & \B{98.3} \\ \hline
         035\_power\_drill       & 3.31 & 69.5  & 74.74 & 51.78 & \B{78.6} \\ \hline
         036\_wood\_block*       & 0    & 2.1   & 2.06  & 90.54 & \B{95.4} \\ \hline
         037\_scissors           & 0    & 12.1  & 56.35 & 60.95 & \B{96.01} \\ \hline
         040\_large\_marker      & 1.38 & 1.9   & 6.8   & 87.92 & \B{93.94} \\ \hline
         051\_large\_clamp*      & 0.28 & 24.2  & 44.94 & 59.31 & \B{86.88} \\ \hline
         052 extra large clamp*  & 0.58 & 1.3   & 7.77  & 21.24 & \B{54.08} \\ \hline
         061\_foam\_brick*       & 0    & 75    & 25    & 67.07 & \B{96.84} \\ \hline \hline
         average                 & 7.25 & 37.99 & 46.18 & 60.29 & \B{87.35} \\ \hline
  \end{tabular}
  }
  \vspace{0.1mm}
  \caption{The accuracy in terms of the ADD(-S) metric (left) and 2D projection error (right) for our method and the baseline methods on the YCB-VIDEO dataset. Symmetrical objects are marked with an asterisk (*) and small objects are marked with $^\dagger$.}
  \label{tab:ycb}
\end{table*}

\vspace{-3mm}

\subsection{6D Object Pose Estimation Evaluation}

\begin{figure*}[!t]
	\centering
	\includegraphics[trim=0 60 0 0, clip, width=0.95\textwidth]{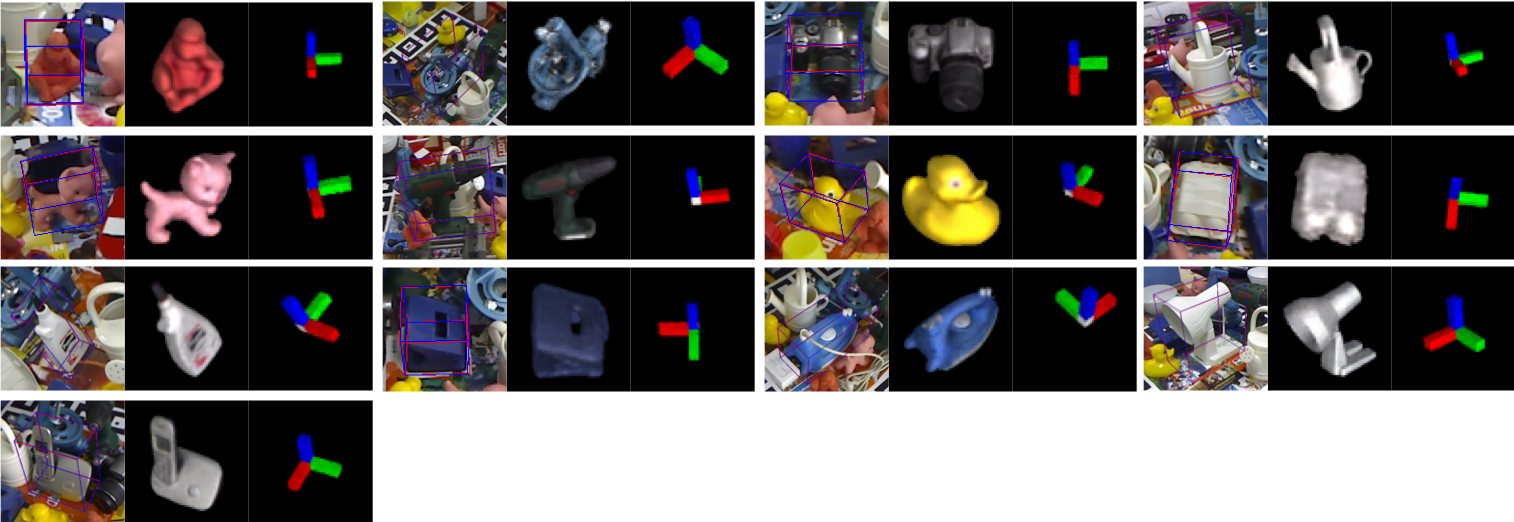}
    \caption{Qualitative pose estimation evaluation on the LINEMOD dataset. The red box indicates the ground truth pose while blue is the estimated pose using PrimA6D. Next to the inferred pose is reconstructed object and the predicted primitive associated to each object.}

	\label{fig:linemod_result_fig}
	\vspace{-4mm}
\end{figure*}

\begin{figure*}[!t]
	\centering
	\includegraphics[width=0.95\textwidth]{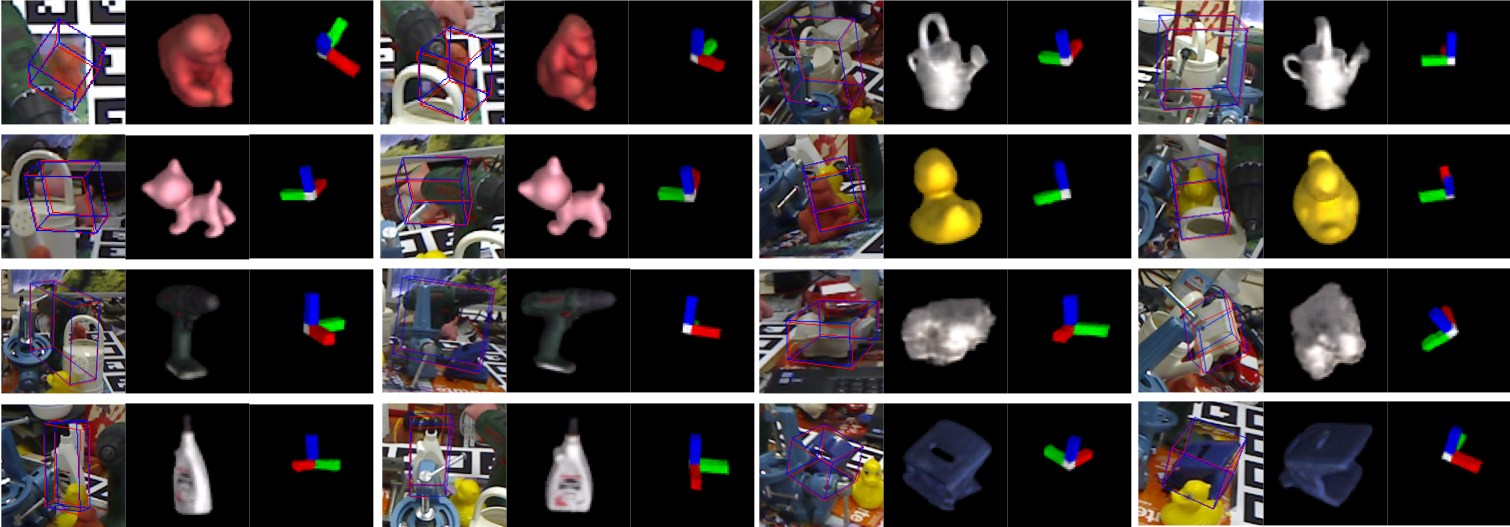}
    \caption{Qualitative pose estimation evaluation on the Occlusion LINEMOD dataset. The red box indicates the ground truth pose while blue is the estimated pose using PrimA6D. Next to the inferred pose is reconstructed object and the predicted primitive associated to each object.}

	\label{fig:linemod_occ_result_fig}
	\vspace{-4mm}
\end{figure*}

We evaluate the performance of our model on the three datasets. The proposed method is compared against other \ac{SOTA} methods from holistic, keypoint-based and part-based approaches. We evaluate our method when training only with synthetic images (PrimA6D-S) and when training using 180 additional real images (PrimA6D-SR). Other existing methods were trained using additional real images. Qualitative results for all three datasets are presented in \figref{fig:linemod_result_fig} and \figref{fig:linemod_occ_result_fig}. Please also refer to \texttt{prima6d.mp4}. We excluded the comparison to the case without the primitive decoder due to drastically large error when no primitive decoder was used.

\subsubsection{LINEMOD Dataset Results}

\tabref{tab:linemod_add} and \tabref{tab:linemod_2dp} exhibit the evaluation for all 13 objects in the LINEMOD dataset. Among the keypoint or the part-based approaches, PVNet is the most recently released method with \ac{SOTA} performance and thus used as the main baseline method for comparison.

Over the LINEMOD dataset, PrimA6D-SR outperformed PVNet in terms of the ADD(-S) score and presented a comparable result in terms of the 2D projection metric. The different performance can be understood from the \ac{MAE} metric in \tabref{tab:linemod_2dp}. If the object center point in the image is found properly, the 2D projection metric is affected more by the rotational estimation accuracy and ADD(-S) is more critical to the translation inference performance. As can be seen, PVNet exhibited higher performance than PrimA6D-S in terms of the rotation \ac{MAE} metric, which results in a performance difference on the 2D projection metric. On the contrary, PrimA6D-S outperformed PVNet in terms of the translation \ac{MAE} metric, which results in a performance difference on the ADD(-S) score.

In addition, the PrimA6D(-S/-SR) presents the strength for small and texture-less objects that the drawback in the PVNet such as ape and cat. Especially, the PrimA6D(-S/-SR) has the forte on the translation estimation.

\subsubsection{Occlusion LINEMOD Dataset Results}

Tables in \tabref{tab:occ_add} list the performance on the Occlusion LINEMOD dataset. Here, we only evaluate the performance of models that do not have the refinement step. As can be seen in two tables, the PrimA6D(-S/-SR) outperformed PVNet in terms of the ADD(-S) score and presented a comparable result in terms of the 2D projection metric. As the PrimA6D(-S/-SR) reconstruct the corresponding object and primitive, we can affirm notable improvement on the occluded case. Moreover, the PrimA6D(-S/-SR) demonstrates striking accuracy for objects that are difficult to recognize by other methods due to small size such as cat and holepuncher.

\subsubsection{YCB-Video Dataset Result}

We further evaluate the pose estimation performance over the YCB dataset as shown in \tabref{tab:ycb}. The PrimA6D(-S/-SR) achieved enhanced inference capability even when trained solely from synthetic dataset (PrimA6D-S). The proposed method is capable of inferring 6D pose of objects that are difficult to recognize by other methods such as wood block.

Another notable point in YCB dataset is the overall down performance from LINEMOD; all approaches including PVNET and ours shows lower accuracy in YCB than LINEMOD. This is because of the challenging and realistic dataset nature of YCB. Despite the challenging dataset, the proposed method shows the smallest performance drop from LINEMOD both in terms of ADD(-S) and 2D projection error even when solely trained from synthetic data. This reveals the generalization capability of the proposed method while others lacking.

\subsection{Ablation Studies}

We further evaluate the proposed method by conducting ablation studies in terms of the effect of real images, effect of training with \ac{GAN} and bounding box accuracy.

\textbf{1) Effect of real image} The proposed method was designed to exploit synthetic data from CAD model. However, considering the potential discrepancy between CAD and real data, additional weight refinement using real training image shows very meaningful performance improvement. We train the network only from synthetic data (PrimA6D-S) and compare the performance improvement by adding real images in the training set (PrimA6D-SR). This comparison is provided for three datasets. Obviously we witness accuracy improvement in all metrics by adding real images in the training set. 

\begin{figure}[!t]
	\centering
	\includegraphics[width=0.95\columnwidth]{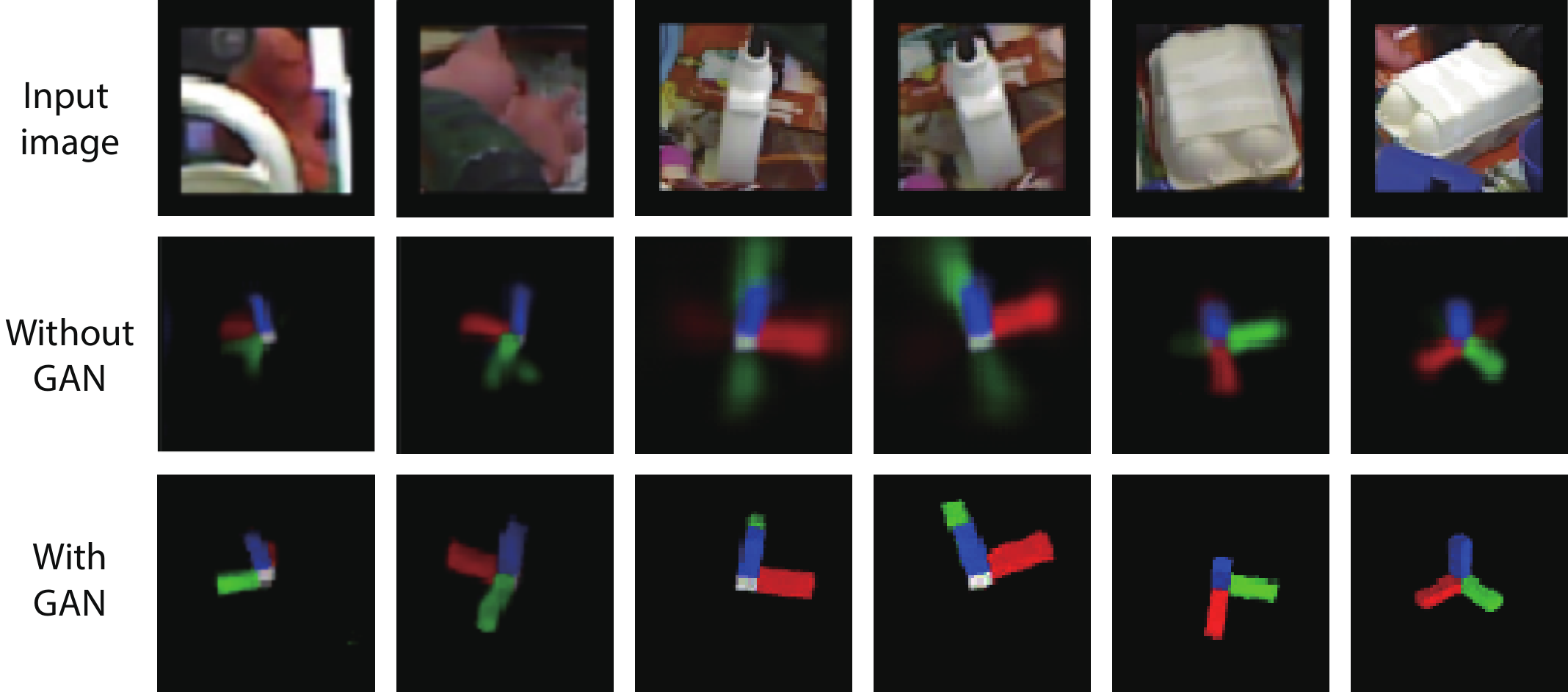}

	\caption{\bl{Effect of GAN on occluded and symmetrical object.}}

	\label{fig:effect_gan}
	\vspace{-4mm}
\end{figure}



\textbf{2) Training with GAN} The adversarial loss and training strategy using \ac{GAN} provides a detailed refinement of the object and primitive reconstruction. As can be seen in \figref{fig:effect_gan}, the reconstructed result shows a blurry and obscure reconstruction when without using \ac{GAN}. This indicates the potential ambiguity in the rotation and translation estimation. Additional training with \ac{GAN} enables the \ac{VAE} to build discriminability. \bl{Also, regarding symmetrical objects with ambiguity for one axis, incorporating adversarial training can solve the potential axis ambiguity in the reconstructed primitive and completely form three orthogonal axes. In the experiment, we confirmed that the ambiguous axis in the primitive becomes the clear orthogonal axis for the other two axes according to the qualitative result.} Testing over Occlusion LINEMOD in \tabref{tab:occ_add} reveals this substantial improvement from the adversarial loss.

\begin{table}[!b]
\resizebox{\linewidth}{!}{
\begin{tabular}{c|ccc|ccc|c}
\hline
method     & \multicolumn{3}{c|}{PrimA6D-S} & \multicolumn{3}{c|}{PrimA6D-SR} & PVNET\\ \hline
IoU        & 1        & 0.9      & 0.75     & 1         & 0.9      & 0.75  & -\\ \hline \hline
LINEMOD    & 87.84    & 85.48    & 77.22    & 97.62     & 97.23    & 90.72 & 86.27 \\ \hline
OCC-LINEMOD & 51.53    & 50.31    & 43.65    & 59.77     & 59.17    & 50.24 & 40.77 \\ \hline
\end{tabular}
}
\caption{Performance variation in terms of the ADD(-S) metric according to \ac{IoU} showing the  accuracy on bounding-box. PVNET results are copied from \tabref{tab:linemod_add} and \tabref{tab:linemod_2dp} for easy comparison.}
\label{tab:bb_ablation}
\end{table}


\textbf{3) Bounding box accuracy} The input of the network is a cropped image from an object detection module hence the PrimA6D estimation performance may be dependent on the detection accuracy. We have examined the effect of the detected bounding box accuracy by reducing the \ac{IoU} to the original bounding box in terms of  ADD(-S) as this is more sensitive to the translational estimation accuracy. As can be seen in \tabref{tab:bb_ablation}, the performance drops as we reduce the \ac{IoU} from 1.0 to 0.75. As can be seen from \tabref{tab:bb_ablation}, \tabref{tab:linemod_add} and \tabref{tab:linemod_2dp}, the PrimA6D-SR still outperforms all other existing methods despite the performance degradation. More notably, the PrimA6D-S shows the exceptional result on the Occlusion LINEMOD dataset even with decreased IoU, being solely trained from synthetic data.

\subsection{Inference Time}

Since our method is applied after object detection, the total inference time is also affected by the running time of the object detection model. For example, we used M2Det \cite{DBLP:journals/corr/abs-1811-04533} that runs at 33.4 \ac{fps}. For the 6D object pose estimation of a single object in an image, the total process takes 31ms on average using a GTX 1080Ti GPU. The entire 6D pose estimation supports 16 to 17 \ac{fps}.

\subsection{Failure Cases}

\begin{figure}[!t]
	\centering
	\includegraphics[width=0.85\columnwidth]{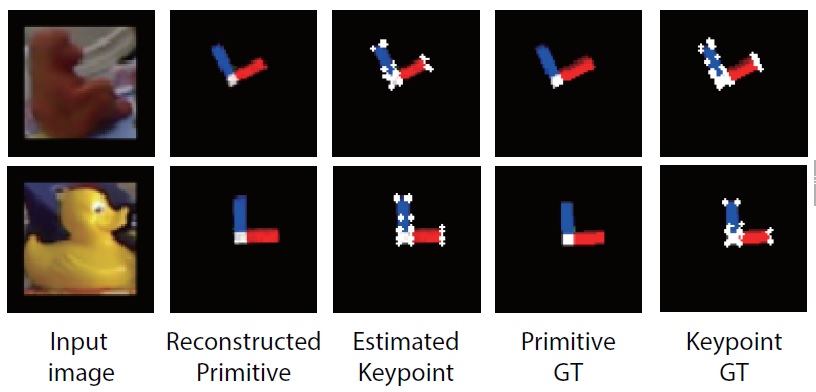}

	\caption{The examples of failure case.}

	\label{fig:fail}
	\vspace{-5mm}
\end{figure}

\figref{fig:fail} shows failure cases when one axis is entirely obscured by the other two axes of the rotation primitive. This hidden primitive axis prevents the algorithm from extracting the correct keypoint yielding a larger rotation error. In future work, we aim to include estimated depth from the primitive to resolve this issue.

%

\section{Conclusion}
\label{sec:conclusion}

This paper reported on 6D pose estimation from a single RGB image by introducing novel primitive learning associated with each object. This paper presents a substantial improvement in the pose estimation even in the occluded case and comparable performance even using only synthetic images for training. The proposed method was validated using three public benchmark datasets yielding \ac{SOTA} performance for occluded and small objects.


\balance
\small
\bibliographystyle{IEEEtranN}
\bibliography{string-short,reference}

\begin{thebibliography}{28}
\providecommand{\natexlab}[1]{#1}
\providecommand{\url}[1]{#1}
\csname url@samestyle\endcsname
\providecommand{\newblock}{\relax}
\providecommand{\bibinfo}[2]{#2}
\providecommand{\BIBentrySTDinterwordspacing}{\spaceskip=0pt\relax}
\providecommand{\BIBentryALTinterwordstretchfactor}{4}
\providecommand{\BIBentryALTinterwordspacing}{\spaceskip=\fontdimen2\font plus
\BIBentryALTinterwordstretchfactor\fontdimen3\font minus
  \fontdimen4\font\relax}
\providecommand{\BIBforeignlanguage}[2]{{%
\expandafter\ifx\csname l@#1\endcsname\relax
\typeout{** WARNING: IEEEtranN.bst: No hyphenation pattern has been}%
\typeout{** loaded for the language `#1'. Using the pattern for}%
\typeout{** the default language instead.}%
\else
\language=\csname l@#1\endcsname
\fi
#2}}
\providecommand{\BIBdecl}{\relax}
\BIBdecl

\bibitem[Peng et~al.(2018)Peng, Liu, Huang, Bao, and Zhou]{pvnet-2019-cvpr}
S.~Peng, Y.~Liu, Q.~Huang, H.~Bao, and X.~Zhou, ``Pvnet: Pixel-wise voting
  network for 6dof pose estimation,'' 12 2018.

\bibitem[Billings and Johnson-Roberson(2019)]{billings2018silhonet}
G.~Billings and M.~Johnson-Roberson, ``Silhonet: An rgb method for 6d object
  pose estimation,'' \emph{{IEEE} Robot. and Automat. Lett.}, 2019.

\bibitem[Periyasamy et~al.(2018)Periyasamy, Schwarz, and
  Behnke]{periyasamy2018robust}
A.~S. Periyasamy, M.~Schwarz, and S.~Behnke, ``Robust 6d object pose estimation
  in cluttered scenes using semantic segmentation and pose regression
  networks,'' in \emph{Proc. {IEEE}/{RSJ} Intl. Conf. on Intell. Robots and
  Sys.}, 2018, pp. 6660--6666.

\bibitem[Wang et~al.(2019)Wang, Xu, Zhu, Mart{\'\i}n-Mart{\'\i}n, Lu, Fei-Fei,
  and Savarese]{wang2019densefusion}
C.~Wang, D.~Xu, Y.~Zhu, R.~Mart{\'\i}n-Mart{\'\i}n, C.~Lu, L.~Fei-Fei, and
  S.~Savarese, ``Densefusion: 6d object pose estimation by iterative dense
  fusion,'' in \emph{Proc. {IEEE} Conf. on Comput. Vision and Pattern Recog.},
  2019, pp. 3343--3352.

\bibitem[Manhardt et~al.(2019)Manhardt, Kehl, and Gaidon]{manhardt2019roi}
F.~Manhardt, W.~Kehl, and A.~Gaidon, ``Roi-10d: Monocular lifting of 2d
  detection to 6d pose and metric shape,'' in \emph{Proc. {IEEE} Conf. on
  Comput. Vision and Pattern Recog.}, 2019, pp. 2069--2078.

\bibitem[{Saxena} et~al.(2009){Saxena}, {Driemeyer}, and
  {Ng}]{saxena-2009-icra}
A.~{Saxena}, J.~{Driemeyer}, and A.~Y. {Ng}, ``Learning 3-d object orientation
  from images,'' in \emph{Proc. {IEEE} Intl. Conf. on Robot. and Automat.},
  2009, pp. 794--800.

\bibitem[Brachmann et~al.(2017)Brachmann, Krull, Nowozin, Shotton, Michel,
  Gumhold, and Rother]{brachmann2017dsac}
E.~Brachmann, A.~Krull, S.~Nowozin, J.~Shotton, F.~Michel, S.~Gumhold, and
  C.~Rother, ``Dsac-differentiable ransac for camera localization,'' in
  \emph{Proc. {IEEE} Conf. on Comput. Vision and Pattern Recog.}, 2017, pp.
  6684--6692.

\bibitem[Brachmann and Rother(2018)]{learning-less-2017}
E.~Brachmann and C.~Rother, ``Learning less is more - 6d camera localization
  via 3d surface regression,'' \emph{2018 IEEE/CVF Conference on Computer
  Vision and Pattern Recognition}, pp. 4654--4662, 2018.

\bibitem[Sundermeyer et~al.(2018)Sundermeyer, Marton, Durner, Brucker, and
  Triebel]{sundermeyer2018implicit}
M.~Sundermeyer, Z.-C. Marton, M.~Durner, M.~Brucker, and R.~Triebel, ``Implicit
  3d orientation learning for 6d object detection from rgb images,'' in
  \emph{Proc. European Conf. on Comput. Vision}, 2018, pp. 699--715.

\bibitem[{Xie} et~al.(2017){Xie}, {Girshick}, {Dollár}, {Tu}, and
  {He}]{DBLP:journals/corr/XieGDTH16}
S.~{Xie}, R.~{Girshick}, P.~{Dollár}, Z.~{Tu}, and K.~{He}, ``Aggregated
  residual transformations for deep neural networks,'' pp. 5987--5995, 2017.

\bibitem[Rad and Lepetit(2017)]{rad2017bb8}
M.~Rad and V.~Lepetit, ``Bb8: A scalable, accurate, robust to partial occlusion
  method for predicting the 3d poses of challenging objects without using
  depth,'' in \emph{Proc. {IEEE} Intl. Conf. on Comput. Vision}, 2017, pp.
  3828--3836.

\bibitem[Tekin et~al.(2018)Tekin, Sinha, and Fua]{tekin2018real}
B.~Tekin, S.~N. Sinha, and P.~Fua, ``Real-time seamless single shot 6d object
  pose prediction,'' in \emph{Proc. {IEEE} Conf. on Comput. Vision and Pattern
  Recog.}, 2018, pp. 292--301.

\bibitem[Brachmann et~al.(2016)Brachmann, Michel, Krull, Ying~Yang, Gumhold,
  et~al.]{brachmann2016uncertainty}
E.~Brachmann, F.~Michel, A.~Krull, M.~Ying~Yang, S.~Gumhold \emph{et~al.},
  ``Uncertainty-driven 6d pose estimation of objects and scenes from a single
  rgb image,'' in \emph{Proc. {IEEE} Conf. on Comput. Vision and Pattern
  Recog.}, 2016, pp. 3364--3372.

\bibitem[Oberweger et~al.(2018)Oberweger, Rad, and
  Lepetit]{oberweger2018making}
M.~Oberweger, M.~Rad, and V.~Lepetit, ``Making deep heatmaps robust to partial
  occlusions for 3d object pose estimation,'' in \emph{Proc. European Conf. on
  Comput. Vision}, 2018, pp. 119--134.

\bibitem[Park et~al.(2019)Park, Patten, and Vincze]{pix2pose}
K.~Park, T.~Patten, and M.~Vincze, ``Pix2pose: Pixel-wise coordinate regression
  of objects for 6d pose estimation,'' 2019, pp. 7667--7676.

\bibitem[Do et~al.(2018)Do, Cai, Pham, and Reid]{do2018deep}
\BIBentryALTinterwordspacing
T.~Do, M.~Cai, T.~Pham, and I.~D. Reid, ``Deep-6dpose: Recovering 6d object
  pose from a single {RGB} image,'' \emph{CoRR}, vol. abs/1802.10367, 2018.
  [Online]. Available: \url{http://arxiv.org/abs/1802.10367}
\BIBentrySTDinterwordspacing

\bibitem[Kehl et~al.(2017)Kehl, Manhardt, Tombari, Ilic, and
  Navab]{kehl2017ssd}
W.~Kehl, F.~Manhardt, F.~Tombari, S.~Ilic, and N.~Navab, ``Ssd-6d: Making
  rgb-based 3d detection and 6d pose estimation great again,'' in \emph{Proc.
  {IEEE} Intl. Conf. on Comput. Vision}, 2017, pp. 1521--1529.

\bibitem[Liu et~al.(2016)Liu, Anguelov, Erhan, Szegedy, Reed, Fu, and
  Berg]{liu2016ssd}
W.~Liu, D.~Anguelov, D.~Erhan, C.~Szegedy, S.~Reed, C.-Y. Fu, and A.~C. Berg,
  ``Ssd: Single shot multibox detector,'' in \emph{Proc. European Conf. on
  Comput. Vision}, 2016, pp. 21--37.

\bibitem[Xiang et~al.(2018)Xiang, Schmidt, Narayanan, and
  Fox]{xiang2017posecnn}
Y.~Xiang, T.~Schmidt, V.~Narayanan, and D.~Fox, ``Posecnn: A convolutional
  neural network for 6d object pose estimation in cluttered scenes,''
  \emph{Robotics: Science and Systems (RSS)}, 2018.

\bibitem[Wu et~al.(2018)Wu, Zhou, Russell, Kee, Wagner, Hebert, Torralba, and
  Johnson]{wu2018real}
J.~Wu, B.~Zhou, R.~Russell, V.~Kee, S.~Wagner, M.~Hebert, A.~Torralba, and
  D.~M. Johnson, ``Real-time object pose estimation with pose interpreter
  networks,'' in \emph{2018 IEEE/RSJ International Conference on Intelligent
  Robots and Systems (IROS)}, 2018, pp. 6798--6805.

\bibitem[Zhao et~al.(2019)Zhao, Sheng, Wang, Tang, Chen, Cai, and
  Ling]{DBLP:journals/corr/abs-1811-04533}
Q.~Zhao, T.~Sheng, Y.~Wang, Z.~Tang, Y.~Chen, L.~Cai, and H.~Ling, ``M2det: A
  single-shot object detector based on multi-level feature pyramid network,''
  \emph{Proceedings of the AAAI Conference on Artificial Intelligence},
  vol.~33, pp. 9259--9266, 07 2019.

\bibitem[Jeon et~al.(2019)Jeon, Lee, Shin, Jang, and Kim]{mjeon-2019-cams}
M.~Jeon, Y.~Lee, Y.-S. Shin, H.~Jang, and A.~Kim, ``Underwater object detection
  and pose estimation using deep learning,'' vol.~52, no.~21.\hskip 1em plus
  0.5em minus 0.4em\relax Elsevier, 2019, pp. 78--81.

\bibitem[Fan et~al.(2017)Fan, Lyu, Ying, and Hu]{atkloss-2017-nips}
Y.~Fan, S.~Lyu, Y.~Ying, and B.~Hu, ``Learning with average top-k loss,'' in
  \emph{Advances in Neural Information Processing Sys. Conf.}, 2017, pp.
  497--505.

\bibitem[Goodfellow et~al.(2014)Goodfellow, Pouget-Abadie, Mirza, Xu,
  Warde-Farley, Ozair, Courville, and Bengio]{goodfellow2014generative}
\BIBentryALTinterwordspacing
I.~Goodfellow, J.~Pouget-Abadie, M.~Mirza, B.~Xu, D.~Warde-Farley, S.~Ozair,
  A.~Courville, and Y.~Bengio, ``Generative adversarial nets,'' pp. 2672--2680,
  2014. [Online]. Available:
  \url{http://papers.nips.cc/paper/5423-generative-adversarial-nets.pdf}
\BIBentrySTDinterwordspacing

\bibitem[He et~al.(2016)He, Zhang, Ren, and Sun]{resnet}
K.~He, X.~Zhang, S.~Ren, and J.~Sun, ``Deep residual learning for image
  recognition,'' \emph{Proc. {IEEE} Conf. on Comput. Vision and Pattern
  Recog.}, pp. 770--778, 2016.

\bibitem[Zakharov et~al.(2019)Zakharov, Shugurov, and Ilic]{zakharov2019dpod}
S.~Zakharov, I.~Shugurov, and S.~Ilic, ``Dpod: 6d pose object detector and
  refiner,'' in \emph{Proceedings of the IEEE International Conference on
  Computer Vision}, 2019, pp. 1941--1950.

\bibitem[Song et~al.(2020)Song, Song, and Huang]{song2020hybridpose}
C.~Song, J.~Song, and Q.~Huang, ``Hybridpose: 6d object pose estimation under
  hybrid representations,'' June 2020.

\bibitem[Hinterstoisser et~al.(2012)Hinterstoisser, Lepetit, Ilic, Holzer,
  Bradski, Konolige, and Navab]{hinterstoisser2012model}
S.~Hinterstoisser, V.~Lepetit, S.~Ilic, S.~Holzer, G.~Bradski, K.~Konolige, and
  N.~Navab, ``Model based training, detection and pose estimation of
  texture-less 3d objects in heavily cluttered scenes,'' 2012, pp. 548--562.

\end{thebibliography}

\end{document}